\documentclass{article}

\usepackage{PRIMEarxiv}

\usepackage[utf8]{inputenc} 
\usepackage{algorithm}
\usepackage{subfigure}
\usepackage{multirow}
\usepackage{algpseudocode}
\usepackage[T1]{fontenc}    
\usepackage{hyperref}       
\usepackage{url}            
\usepackage{booktabs}       
\usepackage{amsfonts}       
\usepackage{nicefrac}       
\usepackage{microtype}      
\usepackage{lipsum}
\usepackage{fancyhdr}       
\usepackage{graphicx}       
\graphicspath{{media/}}     

\title{Leveraging Mixed Precision Quantization to Tackle Gradient Leakage Attacks in Federated Learning}

\author{%
  Pretom Roy Ovi\\
  Department of Information Systems\\
  University of Maryland, Baltimore County\\
  \texttt{povi1@umbc.edu} \\
  \And
  Emon Dey\\
  Department of Information Systems\\
  University of Maryland, Baltimore County\\
  \texttt{edey1@umbc.edu} \\
  \AND
  Nirmalya Roy\\
  Department of Information Systems\\
  University of Maryland, Baltimore County\\
  \texttt{nroy@umbc.edu} \\
  \And
  Aryya Gangopadhyay\\
  Department of Information Systems\\
  University of Maryland, Baltimore County\\
  \texttt{gangopad@umbc.edu} \\
}

\begin{document}
\maketitle

\begin{abstract}
Federated Learning (FL) enables collaborative model building among a large number of participants without the need for explicit data sharing. But this approach shows vulnerabilities when privacy inference attacks are applied to it. In particular, in the event of a gradient leakage attack, which has a higher success rate in retrieving sensitive data from the model gradients, FL models are at higher risk due to the presence of communication in their inherent architecture. The most alarming thing about this gradient leakage attack is that it can be performed in such a covert way that it doesn’t hamper the training performance while the attackers backtrack from the gradients to get information about the raw data. Two of the most common approaches proposed as solutions to this issue are homomorphic encryption and adding noise with differential privacy parameters. These two approaches suffer from two major drawbacks. They are: the key generation process becomes tedious with the increasing number of clients, and noise-based differential privacy suffers from a significant drop in global model accuracy. As a countermeasure, we propose a mixed-precision quantized FL scheme, and we empirically show that both of the issues addressed above can be resolved. In addition, our approach can ensure more robustness as different layers of the deep model are quantized with different precision and quantization modes. We empirically proved the validity of our method with three benchmark datasets and found a minimal accuracy drop in the global model after applying quantization.
\end{abstract}


\section{Introduction}
Federated Learning (FL) has emerged as an alternative to the centralized approach of building a machine learning model, which introduces collaborative training among multiple clients while keeping their dataset private. Although the inherent architecture of FL eliminates the need for explicit data sharing, it still shows vulnerabilities during privacy and robustness attacks. Among the present adversaries in the context of FL, the gradient leakage attack is considered one of the most harmful ones because attackers can successfully reconstitute sensitive training data by secretly snooping on gradient updates during iterative training\cite{zhu2019deep,zhao2020idlg},  all without affecting model training quality.

Two of the most explored prevention methods in the present literature are differential privacy (DP) and homomorphic encryption (HE). First, in the differential privacy approach \cite{wei2021gradient,naseri2020local}, to protect the confidentiality of training sample data, Gaussian or laplacian noise is added with gradients during training. But in this method, the expense of accuracy deteriorates below the threshold level sometimes, while preserving privacy. The second one is homomorphic encryption \cite{aono2017privacy,fang2021privacy}, where key-based encryption is proposed. In this approach, either separate keys are allocated for different participants or the same key for all participants has to be set. But generating different keys for each client elevates the computation complexity. On the contrary, to ensure all clients are getting the same key, we have to enable key sharing among the clients, but such communication is not desirable in FL.

Moreover, researchers have been investigating quantization, typically a model compression scheme to reduce the computation resource requirement of deep models. We point out a new use case of the quantization approach in tackling the gradient leakage attack. In quantization, the gradient values are transitioned into a less precise form according to our choice of bit size. Unless the attacker has some knowledge about the range information of the unquantized gradient, it is highly unlikely to retrieve some sensitive raw data information. We have chosen mixed-precision over single-precision quantization to make our resistance algorithm more robust. This is because, in mixed-precision, the number of iterations to estimate a hyperparameter jumps up to the power of the layer number of the model. Thus, it can make the data extracting process significantly resource exhaustive for the attackers.

Our approach can overcome the disadvantages introduced in DP and HE methods in the sense that, in our approach, it is possible to dequantize the shared quantized gradients at the server end prior to aggregation to maintain almost the same performance level. Since dequantization is not a fully reversible operation, the accuracy may decrease slightly, but it is still superior to differential privacy's accuracy. Also, due to the compressed nature of the quantization, the gradient size required for transmission is less, thus countering the size expansion issue of HE. The specific scientific contributions we offer here are:

\begin{itemize}
\item We conduct a detailed risk assessment in federated learning (FL) scenario due to a gradient leakage attack and propose a quantization-enabled solution to secure a more robust FL framework. Specifically, our approach is built upon the concept of mixed-precision quantization, which is applied to the gradients during the transmission phase.
\item We empirically demonstrate the applicability of our proposed algorithm with three of the most popular FL datasets. In addition, a comprehensive baseline comparison is performed, and we achieve an average 10\% increase in accuracy while keeping the attack success rate to a bare minimum.
\item We present a pertinent ablation study to determine the impact of different hyperparameters used in our federated framework. We find that along with the traits of attack resiliency and accuracy retention, our method can offer another desirable property of reduced communication cost.
\end{itemize}
\section{Related work}
In Federated learning, the server and clients exchange gradients after each round of training. According to \cite{fredrikson2015model,melis2019exploiting}, gradients reveal some properties of the training data. Moreover, recent studies \cite{zhu2019deep,zhao2020idlg,wei2020framework,geiping2020inverting} demonstrated some approaches that can completely steal the training data from gradients with gradient leakage attack. And existing solutions against gradient leakage can be divided into two groups- one is differential privacy \cite{wei2021gradient,geyer2017differentially,naseri2020local} and other one is encryption \cite{fang2021privacy, li2020privacy}.

Differential privacy offers statistical guarantees against the information an attacker can infer from the output of a randomized algorithm. Differential privacy on a client level is feasible, and high model accuracy can be reached when sufficiently many clients (around 1000) are involved \cite{geyer2017differentially}, but specially in cross-silo federated setup, there may not be thousands of clients. DP is effective when each client has access to a large dataset \cite{song2013stochastic}, but FL setup can't guarantee large training datasets for each client. Furthermore, the quantity of accessible data can vary among clients. Authors in \cite{triastcyn2019federated} demonstrated FL with Bayesian DP in a context where data is similarly distributed across participating clients. But FL can't guarantee similar data distribution among all clients. While providing a certain level of differential privacy guarantee, DP limits the performance accuracy of deep learning models. In addition, algorithms require careful privacy parameter selection; otherwise, gradient leakage-induced privacy breach is possible.

The encryption algorithms often used in FL can be broadly classified as Homomorphic Encryption (HE)  \cite{aono2017privacy,fang2021privacy} and Secure Multi-party Computing (SMC) \cite{li2020privacy,liu2020privacy}. While preserving the privacy of training data samples, HE theoretically ensures no performance loss in terms of model convergence \cite{aono2017privacy,fang2021privacy}. However, the effectiveness of HE comes at the expense of computation and memory \cite{rahman2020towards}, which limits its application. Then the data-size of the encrypted models increases linearly with each homomorphic operation \cite{acar2018survey}. Thus, the encrypted models are significantly larger which increases the communication costs. On the other hand, Secure Multi-party Computing (SMC) in FL scenario requires each worker to coordinate with each other during the training process, which is usually impractical.

\section{Methodology}

In this section, we describe the detailed working procedure of our proposed federated learning approach along with integration of mixed precision quantization.

We describe the individual operations carried out on clients (data owner) and server (model owner), the two components of any FL system. Our proposed built on top of the FedAvg\cite{konevcny2016federated} algorithm and some of the notations used in this description are \(\mathcal{N}=\{1, \ldots, N\}\) signify the set of \(N\) clients, each of which has their own dataset \(D_{k \in \mathcal{N}} .\) Each of them trains a local model using their own dataset and only shares the model gradients to the FL server. Then, the global model formation takes place with all the local model gradient  \(\Delta{W}_{G} \) updates which can denoted by \(\Delta{W}=\cup_{k \in \mathcal{N}} \Delta{W}_{k}\). The complete pseudocode of our method is shown in algorithm \ref{algo1} and described below:
\begin{algorithm}[h]
	\caption{Proposed Federated Learning algorithm}
	\begin{algorithmic}[1]
	\Require Clients number $n$ per iteration, local epoches number $E$, and learning rate $\mu$, Local minibatch size $B$, Total number of iteration $T$
	\Ensure Global model $\mathbf{W}_{G}$.
	    \State [Step 1](Server)
	    \State Initialize $\mathbf{w}_{G}^{0}$
	    \State [Step 2](Client)
	    \State \textbf{LocalTraining}$(i, \Delta {W}):$
	    \State Split local dataset $D_{i}$ to minibatches of size $B$ which are included into the set $\mathcal{B}_{i}$.
	    \For{each local epoch $j$ from 1 to $E$}
	        \For{each $b \in \mathcal{B}_{i}$}
	            \State $\mathbf{W} \leftarrow \mathbf{W}-\mu \Delta W(\mathbf{W} ; b)$ 
	            \Comment{($\mu$ is the learning rate, $\Delta W$ is the gradient of $L$ on $b$.)}
	        \EndFor
	    \EndFor
	    \\
	    \textbf{Gradient Quantization:} $\Delta{W_q}\leftarrow Quantize(\Delta{W})$
	    \State[Step 3](Server)\\
	    \textbf{Gradient Dequantization:} $\Delta{W}^{t}\leftarrow
	            Dequantize({\Delta{W}_{q}^{t}})$
	    \State $\Delta{W}_{G}^{t}=\frac{1}{\sum_{i \in \mathcal{N}} D_{i}} \sum_{i=1}^{N} D_{i} \Delta{W}_{i}^{t}$
	    \Comment{(Aggregation through average)}\\
	    \textbf{Updater():}\\
	    \For{each iteration $t$ from 1 to $T$}
	        \State Randomly choose a subset $\mathcal{S}_{t}$ of $m$ clients from $\mathcal{N}$
	            \For{each client $i \in \mathcal{S}_{t}$             \textbf{parallely}}
	                \State $\Delta{W}_{i}^{t+1} \leftarrow$ LocalTraining $\left(i, \Delta{W}_{G}^{t}\right)$
	             \EndFor
	    \EndFor
	    \\
	    \textbf{Gradient Quantization:} $\Delta{W}_{qG}^{t}\leftarrow Quantize(\Delta{W}_{G}^{t})$
	    \State[Step 4](Client)
	    \\
	    \textbf{Gradient Dequantization:} $\Delta{W}_{G}^{t}\leftarrow
	            Dequantize({\Delta{W}_{qG}^{t}})$\\
	    Repeat from step 2.
	\end{algorithmic} 
	\label{algo1}
\end{algorithm} 
\begin{enumerate}
\item \textbf{Executed at the client level}

\begin{itemize}
\item \textit{Local training and update transmission:} Each client at their side uses their own data to learn parameters trained on the received global model during the first training round. The client tries to minimize the loss function \cite{lim2020federated} \(L\left(\Delta{W}_{k}^{t}\right)\) and searches for optimal hyperparameters \(\Delta{W}_{k}^{t}\).
\begin{equation}
\Delta W_k^{t^*}=\Delta W_k^t \arg \min L\left(\Delta W_k^t\right)
\end{equation}
As soon as the training is completed at the clients side, the model gradients are quantized to \(\Delta{W}_{q_G}^{t}\), where \(t\) stands for each iteration index, using mixed-precision algorithm and transmitted to the server. These step prevents the framework from the gradient leakage attack. Even if the gradients are leaked, The permutations required to retrieve the raw data from quantized gradients make the process very arduous for the attackers.
\end{itemize}
\item \textbf{Executed at the server side}
\begin{itemize}
\item \textit{Weight initialization:} The global model \(\mathbf{w}_{G}^{0}\) and it's hyperparameters are disseminated from the server side. In our case, the application we chose is image classification and LeNet-5 model is initiated with randomized weight during the first training round. As soon as the first round finishes, the aggregated gradients achieved from the clients are continued.

\item \textit{Aggregation and global update:} The server first deqauntizes the quantized gradients sent from the participants of each training round. The pseudo code for the dequantization process is given in \ref{algo2}. As soon as the dequantization is done, the aggregation process is carried out to get the aggregated update \(\Delta{W}_{G}^{t+1}\) from the local model gradients of the participants. The server wants to minimize the global loss function \cite{lim2020federated}
\(L\left(\Delta{W}_{G}^{t}\right)\), i.e.
\begin{equation}
    L\left(\Delta{W}_{G}^{t}\right)=\frac{1}{N} \sum_{k=1}^{N} L\left(\Delta{W}_{k}^{t}\right)
\end{equation}
Before sending the updated gradients back to the data owners, the updates are quantizated to \(\Delta{W}_{q_G}^{t+1}\) using the algorithm described in \ref{algo2}.
This process is repeated until the global loss function converges or a desirable training accuracy is achieved. The Global Updater function runs on the SGD \cite{li2020preserving} formula for weight update. The formal equation of global loss minimization formula by the averaging aggregation at the \(t^{t h}\) iteration is given below:
\begin{equation}
\Delta{W}_{G}^{t}=\frac{1}{\sum_{k \in \mathcal{N}} D_{k}} \sum_{k=1}^{N} D_{k} \Delta{W}_{i}^{t}
\end{equation}
\end{itemize}
\begin{algorithm}[h]
	\caption{Quantization and Dequantization}
	\begin{algorithmic}[1]
	\Require $min\_T$ (min value of target tensor), $max\_T$ (max value of target tensor), $max\_float$ (max value of input tensor of type float32)
    	\State [Quantization]
    	\If{($min\_T$ x $min\_range$ > $0$)}
    	\State min\_scale\_factor = $min\_T$ / $min\_range$
    	\Else
    	\State min\_scale\_factor = $max\_float$
    	\EndIf
    	\If{($max\_T$ x $max\_range$ > $0$)}
    	\State max\_scale\_factor = $max\_T$ / $max\_range$
    	\Else
    	\State max\_scale\_factor = $max\_float$
    	\EndIf\\
    	$scale\_factor$ = min(min\_scale\_factor,max\_scale\_factor)\\
    	$min\_range$ = $min\_T$ / $scale\_factor$ \\
        $max\_range$ = $max\_T$ / $scale\_factor$\\
        \textbf{Quantized\_output} = round(min($max\_range$, max($min\_range$, input)) x $scale\_factor$)
        \State [Dequantization]
        \If{(min(T) == $0$)}
    	\State $scale\_factor$ = $max\_range$ / $max\_T$
    	\Else
    	\State $scale\_factor$ = max($min\_range$ / $min\_T$, $max\_range$ / $max\_T$)
    	\EndIf\\
        \textbf{Dequantized\_output} = Quantized\_output * $scale\_factor$
	\end{algorithmic} 
	\label{algo2}
\vspace{-1 mm}
\end{algorithm}
\item \textbf{Quantization and Dequantization}\\
In the method of quantization that we have outlined, scaling factor determination is the key factor during the quantization process. Since we have implemented mixed-precision quantization, the bit size that is used for each layer has a different range of options available to it. The scaling factor is set to have the highest feasible value in order to make sure that all of the values that fall within the range extending from the minimum range to the maximum range are able to be represented by values of our selected data type of output tensor. After that, the scale factor is utilized to make the modifications to the minimum and maximum values. As soon as these steps are completed, quantized version of the input tensor can be obtained by clipping the values to the minimum range and maximum range and then multiplying by the scaling factor.

The dequantization process uses the same set of parameters to select the maximum and minimum range values and looks for the maximum value between the two ranges if the quantized tensor value is not zero. The complete pseudocode for this approach can be found in algorithm \ref{algo2}.
	    
\end{enumerate} 
\section{Experiment and Result Analysis}
\begin{figure}[h]
    \begin{subfigure}
      \centering
      \includegraphics[width=\linewidth]{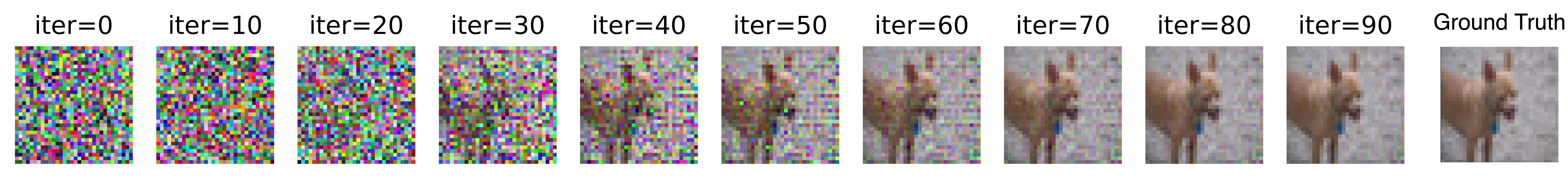}  
      \label{Mnist}
    \end{subfigure}
    \begin{subfigure}
      \centering
      \includegraphics[width=\linewidth]{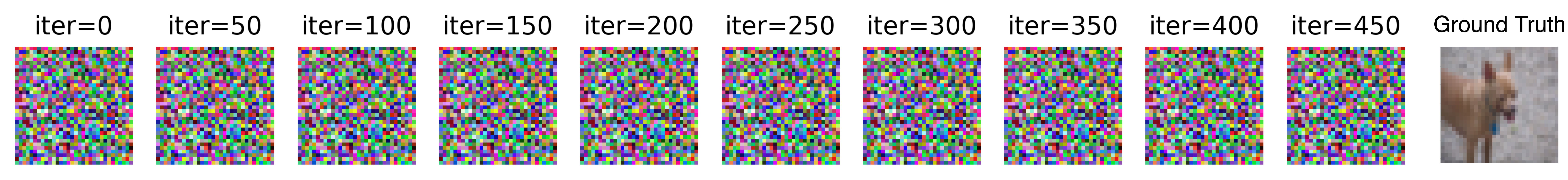}  
      \label{cifar}
    \end{subfigure}
\vspace{-4 mm}
\caption{Gradient leakage attack. Training sample retrieved within 40 iterations (above), no leakage from gradients after mixed precision applied (below).}
\label{entire1}
\vspace{-0.2 cm}
\end{figure}

Server-clients based federated learning framework is implemented on Keras, which is built on TensorFlow backend. The server sets the pace of the training, determines the number of epochs per round, and how many rounds of overall training are to be conducted. In mixed-precision quantization, the gradients of different layers of deep model are quantized with different precision and quantization bits. We utilized the combination of int8 and int16 bit quantization keeping the accuracy in the loop. We conducted the proposed federated training framework with $15$ clients on MNIST \cite{lecun1998mnist}, fashion mnist\cite{xiao2017fashion}, CIFAR-10 \cite{krizhevsky2009learning} dataset.
\begin{figure}[h]
    \begin{subfigure}
      \centering
      \includegraphics[width=\linewidth]{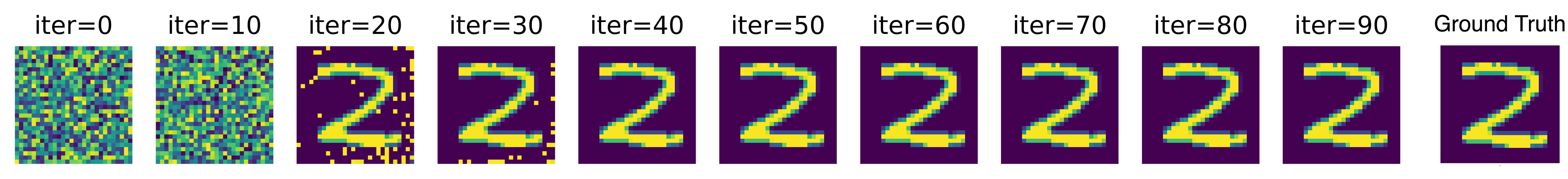}  
      \label{Mnist}
    \end{subfigure}
    \begin{subfigure}
      \centering
      \includegraphics[width=\linewidth]{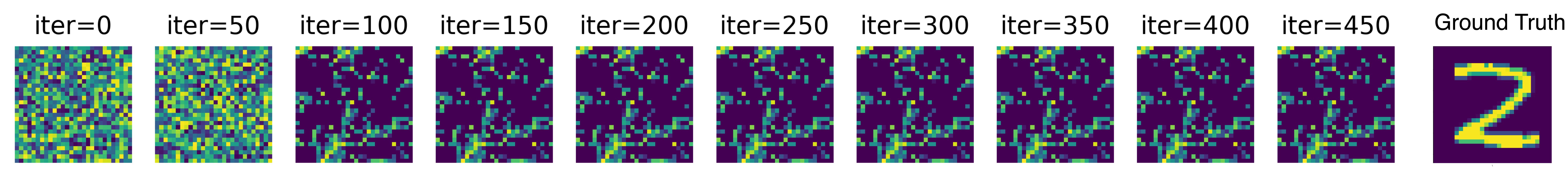}  
      \label{mnist_quant}
    \end{subfigure}
\vspace{-4 mm}
\caption{Gradient leakage attack. Training sample retrieved within 30 iterations (above), no leakage from gradients after mixed precision applied (below).}
\label{entire}
\vspace{-3 mm}
\end{figure}
\\
At first, we launched the gradient leakage attack by extracting the training samples from the gradients. To launch the attack, we attempted to match the gradients generated by the dummy data and the real ones. The difference between the dummy and real data is decreased by minimizing the distance between gradients. Figure \ref{entire} and \ref{entire1} depicted the gradient leaking process that how training samples could be retrieved from gradients. And we found out that recovering monochromatic images with a clean background (MNIST) is easier, whereas recovering relatively complex images (CIFAR-10) requires more iterations.

Secondly, we showed the effectiveness of our proposed mixed precision quantized approach against the gradient leakage attack. From figure \ref{entire} and \ref{entire1}, we can see that training images could not be retrieved from the quantized gradients even after 450 iterations of distance minimization, whereas we retrieved both mnist and cifar-10 samples within 40 iterations from unquantized gradients (figure \ref{entire} and figure \ref{entire1}).
\begin{table}[h]
  \caption{Top-1 accuracy comparison table}
  \label{table}
  \centering
    \begin{tabular}{llll}
        \toprule
        \multirow{2}{*}{} & \multicolumn{3}{c}{Dataset}     \\
        \cmidrule(r){2-4}
                              Methods & Mnist & Fashion Mnist & Cifar10 \\
        \midrule
        Base FL (vulnerable to attack)                & 97.05 & 86.8          & 60.04   \\
        FL with DP (epsilon=1,delta=1e-5)             & 89.94       & 79.49              & 38.39        \\
        Mixed Precision and Quantization (Proposed FL)               & 96.67 & 85.22         & 58.9   \\
        \bottomrule
    \end{tabular}
\end{table}

\vspace{-2 mm}

Moreover, we compared the performance efficiency of our proposed approach with differential privacy based FL. We illustrated the gaussian differential privacy approach with epsilon=$1$ and delta=$1e-5$ as optimum level of noise to provide data privacy. In comparison with base FL (model with no defense against gradient leakage attack), our proposed approach achieved almost the same level of performance for mnist dataset and $1.5$\% degradation for fashion mnist dataset, whereas DP FL degrades around $7$\% on performance for both mnist and fashion mnist. Similarly, DP FL has $21$\% accuracy drop on cifar-10, whereas our approach has only $1$\% drop, demonstrated in table \ref{table}.
\begin{figure}[h]
  \centering
  \includegraphics[width=\linewidth]{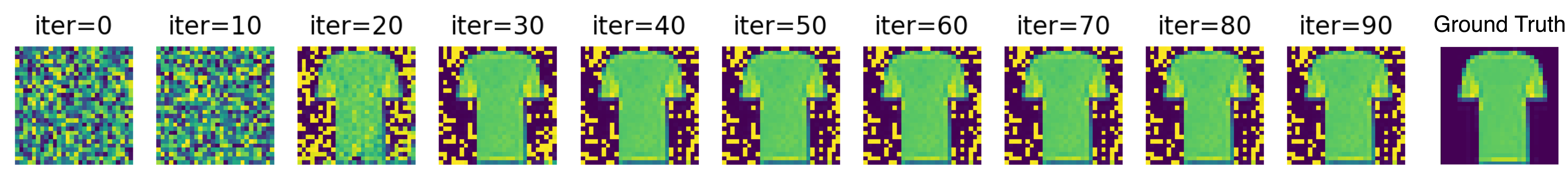} 
  \caption{Recovered image from dequantized gradients with negligible noise in the background.}
  \label{fashion3}
\vspace{-4 mm}
\end{figure}

While ensuring the defense against gradient leakage attacks, the proposed approach has almost same level performance compared to base FL Since we dequantized the quantized gradients at the server side prior to update global model. Since dequantization is not a fully reversible process, the recovered image from dequantized gradients has a few negligible noises in the background pixels (figure \ref{fashion3}) but this process does not necessarily impact the model's performance.

\section{Ablation Study}

\begin{figure}[h]
  \centering
  \includegraphics[width=\linewidth]{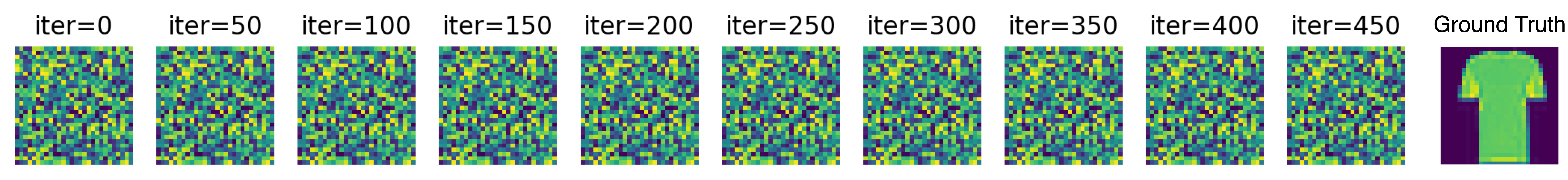}  
  \caption{Failure in image recovery when dequantized mode mismatched with quantized mode}
  \label{fashion2}
\end{figure}

In this section, we analyze our framework from two different points of view.
Firstly, we provide our findings on the impact of specific quantization modes. While implementing our quantized framework, we employ a hyperparameter named 'Quantization mode'. The content determines which calculation procedure will be used to determine the modified maximum and minimum data range. The quantization mode hyperparameter is varied across layers keeping accuracy in the loop. In the server side, the mode that was used for quantization must be selected for dequantization to get the ground truth gradients. Otherwise, dequantized gradients will have a significant mismatch that will largely impact the performance. To showcase the mismatch, we illustrated in figure \ref{fashion2} that ground truth image cannot be recovered even from deqauantized gradients if generated by a different mode. \\

Secondly, the mixed-precision quantization ensures robustness against the attack where an attacker aims to reveal the actual gradient by dequantize the quantized weights. The set of operations that is needed to quantize, the same set of reverse operations is needed to dequantize it. If we assume that the attacker somehow reveals the min max range of float32 bit actual gradients, the attacker may dequantize it by trying all m combinations for single bit quantization where m is the number of quantization mode. On the other hand, for mixed bit quantization, the attacker will need to try $m^L$ combinations to crack it which is a large number of combinations, where L is the number of layers in any deep model.

\section{Discussion}

Our proposed approach can also be seen as a communication efficient federated framework. In our approach, the low bit quantized gradients are being shared between server and clients rather than transmitting the gradients of float32. For instance, converting the precision of activation and gradients from 32-bit floats (174 kB) to 8-bit integers (44 kB) results in $4\times$ data reduction, which eventually requires 4 times less transmission bandwidth. Quantized gradient reduces the downstream and upstream communication cost and thus speeds up training.
\section{Conclusion}

The use of federated learning is motivated by data privacy and communication efficiency. Our proposed mixed precision quantized FL ensures data privacy and low communication cost. Experiments on three benchmark datasets demonstrate that our approach outperforms the differential privacy approach in terms of accuracy and is better than the encryption process in terms of communication cost and computational overheads. To conclude, the proposed federated framework has high resilience against gradient privacy leakage attacks with competitive performance accuracy and a strong data privacy guarantee.

\bibliographystyle{unsrt}  
\bibliography{references}

\end{document}